# Neuromorphic Seatbelt State Detection for In-Cabin Monitoring with Event Cameras

Paul Kielty[1], Cian Ryan[2], Mehdi Sefidgar Dilmaghani[1], Waseem Shariff[1], Joe Lemley[2], and Peter Corcoran[1]

[1]University of Galway, Galway, Ireland
[2]Xperi Corporation, Parkmore Indl. Estate, Galway, Ireland

**Abstract**

Neuromorphic vision sensors, or event cameras, differ from conventional cameras in that they do not capture images at a specified rate. Instead, they asynchronously log local brightness changes at each pixel. As a result, event cameras only record changes in a given scene, and do so with very high temporal resolution, high dynamic range, and low power requirements. Recent research has demonstrated how these characteristics make event cameras extremely practical sensors in driver monitoring systems (DMS), enabling the tracking of high-speed eye motion and blinks. This research provides a proof of concept to expand event-based DMS techniques to include seatbelt state detection. Using an event simulator, a dataset of 108,691 synthetic neuromorphic frames of car occupants was generated from a near-infrared (NIR) dataset, and split into training, validation, and test sets for a seatbelt state detection algorithm based on a recurrent convolutional neural network (CNN). In addition, a smaller set of real event data was collected and reserved for testing. In a binary classification task, the fastened/unfastened frames were identified with an F1 score of 0.989 and 0.944 on the simulated and real test sets respectively. When the problem extended to also classify the action of fastening/unfastening the seatbelt, respective F1 scores of 0.964 and 0.846 were achieved.

## 1 Introduction

Neuromorphic vision describes a class of sensors designed to mimic biological perceptual functions. One such sensor is an event camera, which differs from a conventional camera in that each pixel records data asynchronously. Whenever one of these pixels detects a relative change in brightness above a set threshold an 'event' is logged. Each event is comprised of a timestamp, the coordinate of the pixel that reported the event, and a polarity to indicate whether an increase or decrease in brightness occurred. The event camera does not output images, but a list of events generated by motion or lighting changes in the scene. The event data has no intrinsic framerate, however, its time resolution exceeds that of video captured at 10,000 frames per second. Event cameras also offer higher dynamic range and lower power consumption than most conventional shutter cameras [Gallego et al., 2022].

A 2018 meta-analysis found that a fastened seatbelt reduces the risk of injury in road collisions by 65% [Fouda Mbarga et al., 2018], and in the United States, seatbelt use was shown to reduce mortality by 72% [Crandall et al., 2001]. Existing seatbelt alert systems in modern vehicles rely on pressure sensors in the seat to determine occupancy and simply detect if the seatbelt tongue is inserted in the buckle. This can easily be spoofed by buckling and sitting in front of the seatbelt, and has no ability to determine if a seatbelt has been fastened correctly. Also, it is often only implemented in the front seats of the vehicle. Camera-based seatbelt detection systems have the potential to rectify these flaws. With the ever-increasing demand for safer, more intelligent vehicles, there have been remarkable developments in camera-based DMS. At this stage they have been fully implemented in many modern consumer vehicles. With the camera systems already in place, it is possible to add new DMS features with minimal additional cost. Recent research has revealed how event cameras hold many advantages over standard shutter cameras in for driver monitoring tasks, particularly when it comes to face and eye motion analysis [Ryan et al., 2021, Chen et al., 2020]. In this paper, we demonstrate the viability of another feature in an event-based DMS by creating the first event-based seatbelt state detector.

## 2 Event Data Simulation and Collection

An obstacle regularly faced in event camera research is the lack of publicly available large-scale datasets. This has driven the development of event simulators such as V2E [Delbrück et al., 2020], which enables the synthesis of realistic events from NIR or RGB videos by analysing the differences between consecutive frames. Most of the event data used for this research was simulated with V2E from a non-public industry dataset of NIR videos. Using a wide field of view camera on the rear-view mirror of a car, various subjects were recorded fastening and unfastening their seatbelts repeatedly. The video frames were labelled according to the following classes: (0) The subject's seatbelt is fastened. (1) The subject's seatbelt is unfastened. (2) The subject is fastening their seatbelt. (3) The subject is unfastening their seatbelt. A set of real event data was also collected for testing the network. A Prophesee EVK4 event camera was mounted beside the rear-view mirror of a driving simulator and focused on the driver's seat. Six subjects were asked to fasten and unfasten their seatbelt at random intervals throughout each recording. These videos were labelled manually with the same 4 classes as the NIR dataset.

## 3 Pre-processing of Event Data

The event data, both simulated and real, are saved as lists of events in text format. To use this data in CNNs and other image-based systems, it must first be represented in a 2D array. This is typically achieved by accumulating a group of events and summing the positive and negative events at each pixel location to create a 2D frame [Gallego et al., 2022]. When transforming an event recording into frames with this technique, the decision of how many events should be accumulated per frame must be carefully considered. The two most common approaches are to accumulate events over a fixed duration or accumulate a fixed number of events for each frame. The former method of grouping the events by a fixed duration is useful in tasks that could benefit from the temporal information in a sequence of frames as the generated frames will have fixed time spacing, much like conventional video formats. However, this approach is prone to generating frames with few events if there is little motion in the scene over the fixed duration. The alternative approach of forming each frame from a fixed number of events gives some assurance of a minimum amount of spatial information in each frame, at the loss of much of this temporal information. This works better for keeping the seatbelt visible when there is minimal motion in the frame, but motion of the head or background can quickly saturate the event count and generate many frames where the seatbelt is absent. A custom accumulation approach was developed for the final iteration of the dataset. Each frame was defined by a fixed number of events, however, only events within a rectangle bounding the subject's torso were counted. This maintained seatbelt visibility in more frames than the original two methods, as demonstrated in Fig. 1. where the fixed counts/duration were specified so each method generates 75 frames of the same "Seatbelt Fastened" clip. In the full 75 frames, the seatbelt was visible in (a) 27%, (b) 71%, and (c) 93%. The final dataset contained 108,691 synthetic event frames and 8,317 real event frames. The simulated videos were randomly separated into training, validation, and test sets. The real event videos were all reserved for the testing.

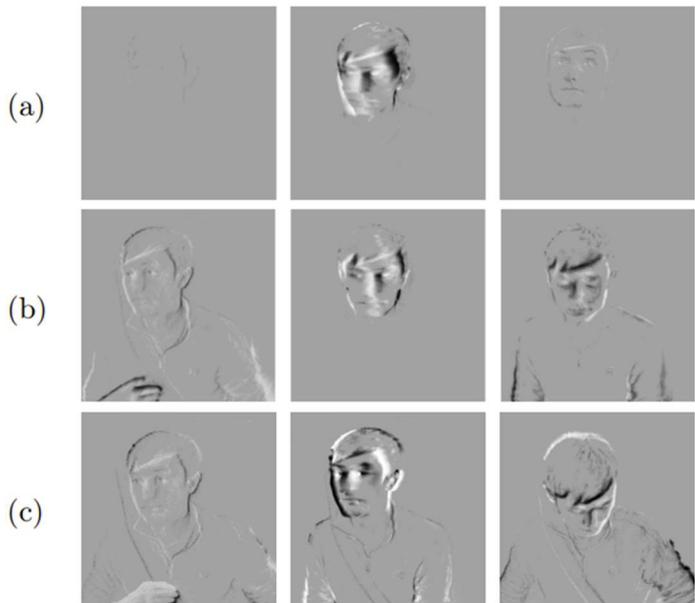

Figure 1. Frames from a "Seatbelt Fastened" event clip generated using (a) a fixed time period, (b) a fixed event count, and (c) a fixed event count over the torso region.

## 4 Network Architecture

It is difficult to distinguish individual fastening/unfastening frames, but it becomes obvious when the whole sequence of frames is considered. Additionally, for the static classes with unreliable seatbelt visibility, using a sequence of frames can provide a more reliable result. For these reasons we used a recurrent CNN architecture which takes a frame sequence is as the input for each prediction. Fig. 2 gives a high-level overview of the structure. The MobileNetV2 network is used as an efficient, lightweight backbone for initial feature extraction [Sandler et al., 2018]. Recent years have seen self-attention introduced to many CNN tasks for its ability to contextualize and apply a weighting to input features, with only a small computational cost. The self-attention module in our proposed network is implemented according to [Zhang et al., 2018]. When attended feature maps have been generated for every frame of the input sequence, they are stacked and passed to the recurrent head of the network. This is comprised of a 2 stacked bi-directional LSTM layers [Hochreiter and Schmidhuber, 1997].

## 5 Training

In this work, two models were trained. The first was for binary classification of frame sequences using the static fastened/unfastened classes only. For the second model, all classes were included to determine if the 4 states could be reliably identified, as they must all be handled in a real-world implementation. To train the network, the videos were split into single-class sequences of 15 frames, before randomized cropping and downsampling to a resolution of 256x256. Using cross-entropy loss and a batch size of 15 sequences, the network was trained for 30 epochs. The initial learning rate of $1 \times 10^{-4}$ was halved every 5 epochs.

An added benefit of the self-attention layer is allowing us to visualize the areas in each frame that are more heavily weighted by the network. This can be helpful to verify that the network is utilizing appropriate features. Fig. 3 shows these weighted regions tracking the seatbelt when visualized on the real event videos in the test set.

## 5 Results and Conclusion

The results of the 2-class model and 4-class model on both the simulated and real test sets are compared in Table 1. As expected, the

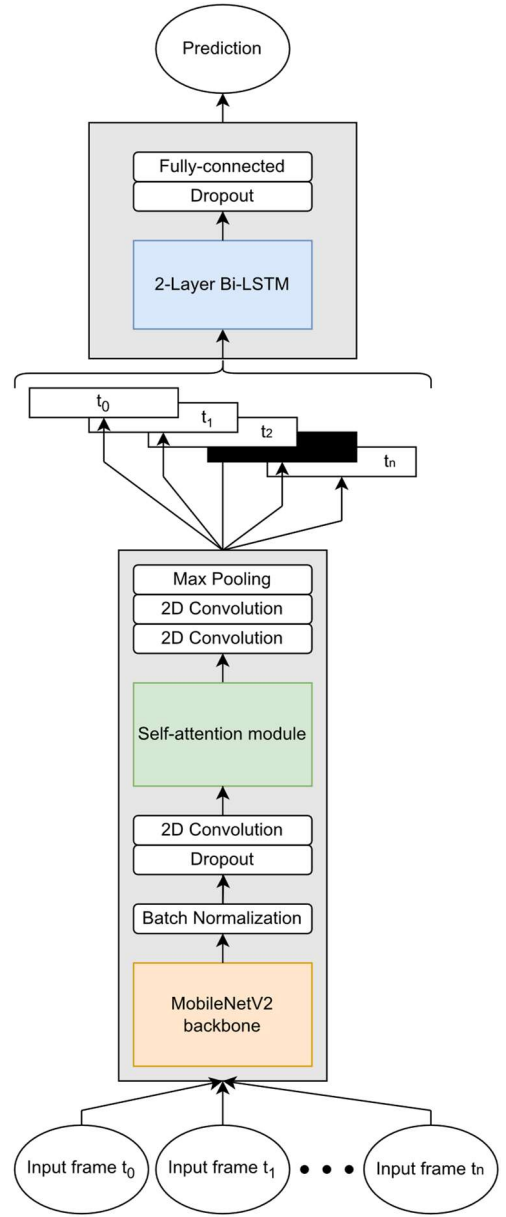

**Figure 2: Proposed network architecture.**

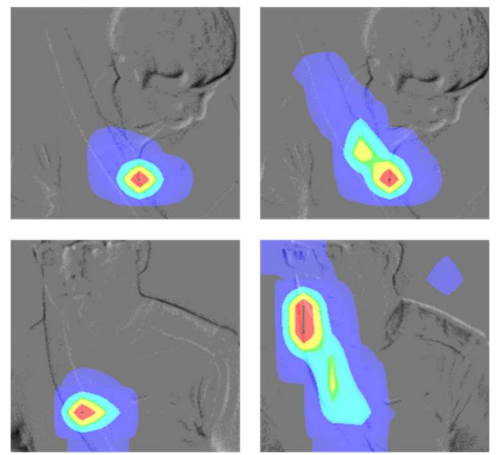

**Figure 3. Visualized attention maps on test frames generated from real events.**

2-class model was more accurate, but the 4-class model demonstrates that handling of all classes is possible without a dramatic reduction in performance. This model treats the 4 classes as independent, but we know they can only transition in a fixed sequence. Future work will leverage this fact for improved accuracy.

| Model | Test set | F1 |
|---|---|---|
| 2-class | Simulated | 0.989 |
|  | Real | 0.944 |
| 4-class | Simulated | 0.964 |
|  | Real | 0.846 |

Table 1. Summary of model performance.

## Acknowledgements

This research was conducted with the financial support of Science Foundation Ireland at ADAPT, the SFI Research Centre for AI-Driven Digital Content Technology at the University of Galway [13/RC/2106_P2]. For the purpose of Open Access, the author has applied a CC BY public copyright licence to any Author Accepted Manuscript version arising from this submission.## References